\documentclass{article}

\usepackage{arxiv}

\usepackage[utf8]{inputenc} 
\usepackage[T1]{fontenc}    
\usepackage{hyperref}       
\usepackage{url}            
\usepackage{booktabs}       
\usepackage{amsfonts}       
\usepackage{nicefrac}       
\usepackage{microtype}      
\usepackage{lipsum}
\usepackage{graphicx}
\usepackage{natbib}

\usepackage{algorithm}
\usepackage{algpseudocode}
\usepackage{comment}
\usepackage{amssymb}
\usepackage{array}

\graphicspath{ {./images/} }

\title{ZapGPT: Free-form Language Prompting for Simulated Cellular Control.}

\author{
 Nam H. Le \\
  Department of Computer Science\\
  University of Vermont\\
  Burlington, VT 05405, USA \\
  \texttt{nam.le@uvm.edu} \\
  \And
  Patrick Erickson \\
  Department of Biology\\
  Tufts University\\
  Boston, MA 02155, USA \\
  \texttt{Patrick.Erickson@tufts.edu }
  \And
  Yanbo Zhang \\
  Department of Biology\\
  Tufts University\\
  Boston, MA 02155, USA \\
  \texttt{Yanbo.Zhang@tufts.edu }
   \And
 Michael Levin \\
  Department of Biology\\
  Tufts University\\
  Boston, MA 02155, USA \\
  \texttt{michael.levin@tufts.edu} \\
  \And
 Josh Bongard \\
  Department of Computer Science\\
  University of Vermont\\
  Burlington, VT 05405, USA \\
  \texttt{jbongard@uvm.edu} \\
}

\begin{document}
\maketitle
\begin{abstract}
Human language is one of the most expressive tools for conveying intent, yet most artificial or biological systems lack mechanisms to interpret or respond meaningfully to it. Bridging this gap could enable more natural forms of control over complex, decentralized systems. In AI and artificial life, recent work explores how language can specify high-level goals, but most systems still depend on engineered rewards, task-specific supervision, or rigid command sets, limiting generalization to novel instructions. Similar constraints apply in synthetic biology and bioengineering, where the locus of control is often genomic rather than environmental perturbation. A key open question is whether artificial or biological collectives can be guided by free-form natural language alone, without task-specific tuning or carefully tuned evaluation metrics. We provide one possible answer here by showing, for the first time, that simple agents’ collective behavior can be guided by free-form language prompts: one AI model transforms an imperative prompt into an intervention that is applied to simulated cells; a second AI model scores how well the prompt describes the resulting cellular dynamics; and the former AI model is evolved to improve the scores generated by the latter. Unlike previous work, our method does not require engineered fitness functions or domain-specific prompt design. We show that the evolved system generalizes to unseen prompts without retraining. By treating natural language as a control layer, the system suggests a future in which spoken or written prompts could direct computational, robotic, or biological systems to desired behaviors. This work provides a concrete step toward this vision of AI/biology partnerships, in which language replaces mathematical objective functions, fixed rules, and domain-specific programming.
\end{abstract}

{\raggedright Data/Code available at: \url{https://github.com/selfemergence/ALife_ZapGPT}\par}


\section{Introduction}
Decentralized, dynamical systems, such as multicellular collectives or agent-based simulations, exhibit rich behavior, but are difficult to steer using intuitive, high-level intents (\cite{camazine2001self, bedau2000open}). While human natural language offers a flexible and expressive channel for conveying these goals, most such systems lack any capacity to interpret or respond meaningfully to it. Enabling lifelike systems to be shaped directly by natural language could radically expand our ability to interact with them, unlocking applications in artificial morphogenesis \cite{levin2021bioelectric}, collective robotics \cite{rubenstein2014programmable}, and programmable synthetic intelligence \cite{sole2016synthetic}.

In AI and robotics, natural language is increasingly used to guide agent behavior, but most implementations still rely on rigid command templates or engineered reward functions (\cite{matthews2019word2vec, zeng2023large}). Meanwhile, in fields like synthetic biology and morphogenetic engineering, control is typically applied at the molecular or genomic level, through gene circuits or chemical gradients, rather than through high-level, interpretable interfaces (\cite{levin2021bioelectric, pezzulo2016top}). Despite rapid progress in both domains, a general-purpose interface that allows decentralized systems to flexibly interpret and respond to free-form linguistic input remains elusive. 

Recent advances in foundation models have demonstrated the ability to map natural language prompts to diverse modalities, including protein structures \cite{Liu2025}, molecular graphs \cite{wu2023molformer}, and 3D spatial forms \cite{poole2022dreamfusion}. These systems enable natural language to serve as a high-level interface to complex, structured outputs, suggesting new ways to interact with dynamical processes. While such models have begun transforming fields like molecular biology and graphics, their integration into Artificial Life is still in its infancy. One recent effort by \cite{kumar2024automating} applied a vision-language model (CLIP; \cite{radford2021learning}) to evaluate ALife simulation outcomes against handcrafted textual prompts, enabling automated search over simulation spaces that resemble desired concepts (e.g., “a caterpillar” or “viral colony”). Although promising, such approaches still rely on domain-specific prompt engineering, evaluation heuristics, and limited language expression. These developments raise a new question: can collective, agent-based systems be guided using only free-form natural language, without engineered reward functions or domain-specific knowledge?

Here we show for the first time that a language-driven framework, which we call ZapGPT, can enable simulated cells to respond to open-ended natural language. The system consists of two components: a Prompt-to-Intervention (P2I) model that maps sentence embeddings to spatial vector fields which influence cell dynamics, and a pre-trained vision-language model (VLM) that evaluates how well the final system behavior aligns with the input prompt. The P2I model is optimized via evolutionary strategies to maximize this alignment score, closing the loop between linguistic intent and emergent behavior.  Although the system is trained solely on the prompt ``form a cluster,'' it is able to produce plausible responses to a range of unseen prompts, including those with both similar and opposite meanings. These initial results suggest that the model captures some degree of semantic grounding, opening a path toward language-based modulation of decentralized behavior without task-specific tuning.

\section{Literature Review}

Complex systems, ranging from biological collectives to artificial agent-based models, exhibit emergent behavior arising from simple, local interactions among components, often producing non-linear and unpredictable global dynamics \citep{camazine2001self, mitchell2009complexity}. The field of Artificial Life (ALife) has long taken interest in such decentralized systems, investigating how lifelike properties and behaviors can arise without centralized control \citep{langton1997artificial, bedau2000open}. Early efforts largely emphasized understanding emergence for its own sake—revealing how simple rule sets can give rise to complexity. However, a growing body of work has shifted attention toward \textbf{goal-directed emergence}, exploring how system-level constraints can be applied to influence or steer emergent outcomes \citep{deacon2011incomplete, winning2019being}. This direction is echoed in regenerative biology and morphogenetic engineering, where researchers such as \cite{Levin2012} and \cite{Adams2012} demonstrate how bioelectric fields and environmental gradients can guide development and repair in a distributed but coordinated fashion—effectively imposing top-down control without micromanaging individual components.

Traditional approaches in synthetic biology and bioengineering, by contrast, have largely focused on bottom-up control: directly modifying genetic pathways, engineering molecular interactions, or constructing logic-based gene circuits to elicit desired behaviors \citep{Elowitz2000, Cameron2014}. While powerful, these methods often lack adaptability and require extensive domain knowledge. In response, alternative strategies have emerged. For instance, bioelectric signaling has been increasingly recognized as a powerful, reprogrammable layer of control over tissue patterning and morphogenesis—allowing spatial organization to be regulated at higher levels of abstraction \citep{levin2021bioelectric, pezzulo2016top}. Despite their promise, however, these mechanisms still require manual specification of intervention parameters, and do not support semantic or intuitive forms of control. The challenge remains: how can complex biological collectives be guided through interfaces that are expressive, general-purpose, and interpretable?

Recent advances in natural language processing (NLP) and large language models (LLMs) have enabled AI systems to interpret and generate human-like language fluently (e.g., GPT \citep{radford2018improving}, BERT \citep{devlin2019bert}). These models have been applied to control robots \citep{zeng2023large, yu2023language, saycan2022arxiv}, interact with virtual agents \citep{thoppilan2022lamda}, generate executable code from language prompts \citep{chen2021evaluating}, and support autonomous decision-making \citep{shinn2023reflexion}.

Research in embodied AI and language grounding explores how linguistic input can be mapped to sensorimotor actions in embodied agents. For example, SayCan \citep{saycan2022arxiv} pairs large language models with value functions to select robotic actions based on natural language goals. RT-2 \citep{brohan2023rt} further demonstrates that vision-language-action models trained on web-scale data can generalize across robotic tasks. Lang2Reward \citep{yu2023language} integrates language with reinforcement learning to generate low-level control policies conditioned on user instruction. While these systems achieve impressive results, they often operate within constrained task spaces or require task-specific scaffolding.

Despite growing interest in using language for control in AI, applications within ALife remain limited. \cite{nisioti2024text} has highlighted this untouched potential for integrating LLMs into ALife,  advocating for deeper, reciprocal connections between language systems and lifelike simulations. Their position paper surveys early attempts to use LLMs for mutation guidance, generative narratives, and steering open-ended evolution, but notes that ALife has yet to fully explore the use of language as an interface for decentralized, emergent systems. Another recent contribution is LifeGPT by \cite{berkovich2024lifegpt}, which proposes a transformer-based architecture for learning the dynamics of cellular automata from raw trajectories. While their approach showcases the representational power of pretrained sequence models in capturing localized update rules, it does not address language-based control or interpretability, focusing instead on unsupervised prediction of state transitions in spatial systems. 

In parallel, \cite{kumar2024automating} introduced a framework, named \textit{ASAL}, that uses CLIP to evaluate whether ALife simulations produce images aligned with textual prompts. While their system enables automated exploration of simulation parameters across platforms such as Boids, Lenia, and cellular automata, it does not dynamically guide behavior. Instead, it performs post-hoc search over parameter settings that yield visually descriptive outcomes, scored against fixed prompts. Although CLIP offers a simple and powerful method for aligning likelihood of texts and images, it cannot generate descriptive feedback in open-ended form; instead, it requires handcrafted prompt engineering by the experimenter, often informed by prior domain knowledge of the simulation outputs.

One recent attempt to link language and decentralized behavior control is \cite{le2025zapgpt}, which introduced a closed-loop framework combining a prompt-to-intervention model with a vision-language evaluator. While demonstrating initial feasibility, the system operated under narrow constraints: both input prompts and evaluation targets were restricted to single-word tokens (e.g., ``cluster''), and scoring was based on handcrafted comparisons with a fixed vocabulary (e.g., ``scattered'' vs. ``clustered''). This design framed evaluation as a binary classification task, which limited semantic expressiveness and blocked open-ended feedback or generalization. As the authors noted, the system failed to respond appropriately to prompts beyond those encountered during training—highlighting a key challenge in grounding language for decentralized control.

\textit{Here, we extend this line of work by introducing ZapGPT}, a system capable of interpreting and responding to \textit{free-form} natural language prompts without prompt engineering or retraining. Our framework supports arbitrary sentence-level instructions and uses a pretrained vision-language model to evaluate behavior based on open-ended, semantic alignment with the original prompt, rather than keyword matching.

\section{Model Description}

\subsection{Overview}

Our system follows a six-stage pipeline (Figure \ref{fig:zapgpt_pipeline}) that maps natural language instructions into agent behavior via a P2I model and a vision-language feedback loop. Below we briefly describe each step:

\begin{figure}[ht]
    \centering
    \includegraphics[width=\linewidth]{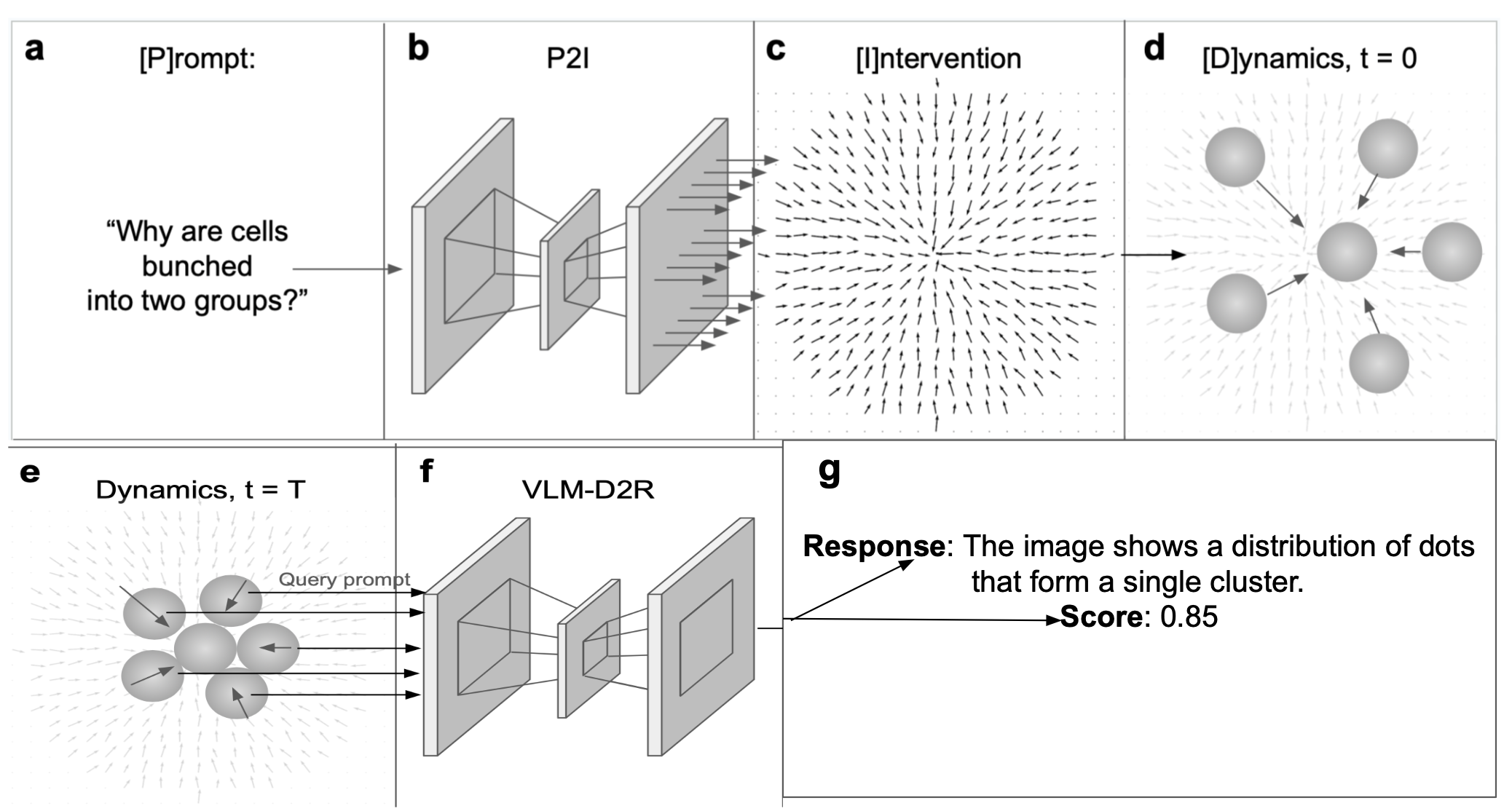}
    \caption{
        Overview of the ZapGPT pipeline. A natural language prompt (a) is converted into a spatial vector field by the P2I model (b–c), driving cellular dynamics in a physics-based environment (d–e). The final configuration is evaluated using a vision-language model (VLM-D2R; f), which produces both a natural language description and a scalar alignment score used for optimization.
    }
    \label{fig:zapgpt_pipeline}
\end{figure}

\subsubsection{(a) Prompt – Natural Language Instruction}

The system begins with a user-provided, free-form natural language prompt (e.g., \textit{``form a cluster''}), which conveys high-level intent without constraints on syntax or vocabulary. Prompts may vary in phrasing (e.g., \textit{``gather the agents into a group''}, \textit{``assemble the cells''}) but express similar goals. This instruction is the sole user input and is passed to the Prompt-to-Intervention (P2I) model, which translates linguistic intent into spatial control signals that guide cellular behavior.

\subsubsection{(b) Prompt-to-Intervention – P2I Model}

\begin{figure}[h]
\centering
\includegraphics[width=\linewidth]{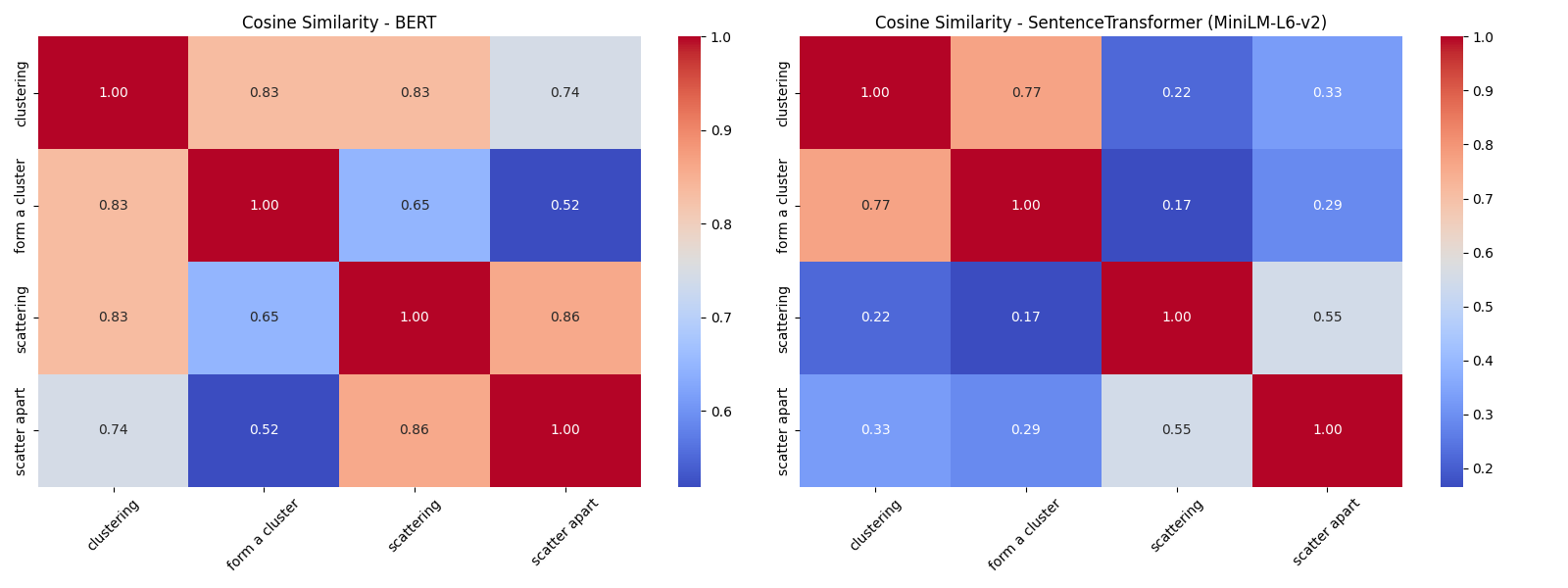}
\caption{
Cosine similarity between prompt embeddings using BERT (left) vs. SentenceTransformer (right). While BERT yields uniformly high similarity, even for semantically opposite prompts, SentenceTransformer produces more discriminative embeddings that better reflect the intent expressed in language.
}
\label{fig:similarity_heatmaps}
\end{figure}

\begin{figure}[h]
    \centering
    \includegraphics[width=0.8\textwidth]{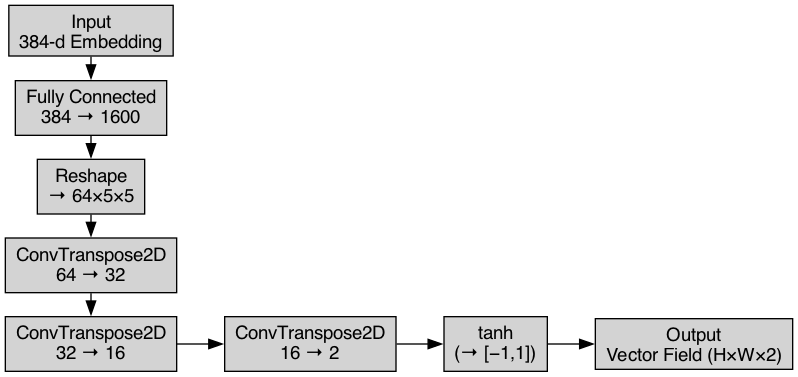}
    \caption{P2I architecture. A 384-dimensional prompt embedding is projected into a $5 \times 5 \times 64$ latent representation, then decoded into a $H \times W \times 2$ spatial vector field via transposed convolutions.}
    \label{fig:cnn_p2i_arch}
\end{figure}

The P2I module maps a user’s natural language prompt into a spatial vector field that serves as the system’s core intervention. The architecture is designed to enhance both semantic expressiveness and spatial coherence, and features: (1) a SentenceTransformer (SBERT)~\citep{reimers-2019-sentence-bert} encoder that generates discriminative prompt embeddings capable of capturing subtle phrasal nuances (e.g., differentiating ``cluster'' from ``scatter'', see Figure \ref{fig:similarity_heatmaps} for cosine similarity comparison between 2 embedders); and (2) a convolutional decoder that introduces local spatial correlations and parameter sharing. The SBERT embedding is projected into a $5 \times 5 \times 64$ latent representation and subsequently upsampled into a $H \times W \times 2$ spatial vector field via transposed convolutions (Figure~\ref{fig:cnn_p2i_arch}). This structured decoding process produces smoother, spatially coherent interventions and supports generalization to a broader range of linguistic inputs.

\subsubsection{(c) Intervention – Spatial Vector Field}

The P2I module outputs a low-resolution 2D vector field (e.g., $5 \times 5$ or $10 \times 10$), where each vector represents a directional force for a region of the environment. Interpolated across space and applied uniformly at each timestep, this field nudges agents based on the prompt’s intent. Unlike engineered controllers, these spatial interventions are learned directly from vision-language feedback and require no task-specific rules or metrics. The result is an abstract but structured link between symbolic language and emergent physical behavior.

\subsubsection{(d-e) Dynamics – Agent Simulation}

The vector field generated by the P2I is applied to a decentralized system of agents (cells), each with simple local behaviors. At the initial timestep ($t = 0$), 50 circular agents are randomly placed in a $500 \times 500$ pixel environment. Each agent experiences two types of forces: repulsion from nearby agents to avoid overlap, and external directional influence from the interpolated vector field.

As the simulation progresses over $T$ timesteps (typically $T = 500$), agents update their velocities and positions in continuous space. The emergent behavior arises from the interaction of the global intervention with local, physics-inspired rules. Agents do not have access to the prompt or its embedding; they respond only to the external field and one another.

At the final timestep ($t = T$), the environment reaches a terminal configuration that reflects the influence of the linguistic prompt. This final state is captured as a rendered image and passed to the semantic evaluation module (D2R).

\subsubsection{(f) VLM-D2R – Vision-Based Scoring}

The D2R module evaluates the final system behavior using a vision-language model (VLM). It takes the rendered simulation image and the original prompt as input, and returns both a natural language description and a scalar alignment score, referred to as the \textit{P2I–D2R alignment score}.

Unlike the original ZapGPT~\citep{le2025zapgpt}, which relied on domain-specific metrics and a fixed prompt set, we use an open-ended evaluator: \textbf{Mistral-Vision}~\citep{mistralvision2024}. This model describes the observed outcome and rates how well it aligns with the user's intent.

We frame the evaluation query as:
{\small
\begin{quote}
Briefly describe the overall distribution of dots in this image. \\
The original prompt was: “\{prompt\}.” \\
On a scale from 0.0 to 1.0, how well does the image match the user’s goal? \\
Return a numerical score.
\end{quote}
}

This scoring approach eliminates the need for handcrafted heuristics, enabling both specification and evaluation to operate entirely in natural language.

\subsubsection{(g) Output – Response and Score}
The final step yields two outputs from the D2R evaluator: a free-form textual description (e.g., \textit{``the agents form a tight cluster near the center''}) and a scalar alignment score between 0.0 and 1.0 indicating how well the behavior matches the original prompt. This score serves as the fitness signal during training, higher values indicate better semantic alignment between intent and outcome. No handcrafted reward functions or task-specific metrics are required; evaluation is performed entirely through language. These outputs close the loop between symbolic instruction and emergent behavior, enabling the system to evolve meaningful responses without explicit programming or domain modeling.

\subsection{Evolutionary Optimization}

Because the simulator is non-differentiable and the D2R alignment score is only available at the final timestep, we use a $(\mu + \lambda)$ evolution strategy \citep{beyer2002evolution} to optimize the weights of the P2I model. Each individual in the population represents a set of weights that maps a prompt to a spatial vector field. For each generation, the system simulates agent behavior under this field, renders the final state, and queries the vision-language model (VLM) for an alignment score, which serves as the fitness signal. Offspring are produced via Gaussian mutation and real-valued crossover, and the best-performing individuals are selected to form the next generation. This approach does not require handcrafted rewards or differentiable environments, and enables generalization to new prompts without retraining. The training loop is summarized in Algorithm~\ref{alg:mu-lambda}.

\begin{algorithm}[H]
\caption{Evolve P2I with VLM-based Fitness}
\label{alg:mu-lambda}
\begin{algorithmic}[1]
\small
\State \textbf{Input:} Prompt $p$, Text Encoder, VLM, population size $\mu$, offspring size $\lambda$, generations $G$
\State Encode prompt $p$ into embedding vector $e$
\State Initialize $\mu$ random models as population $P$
\For{$g = 1$ to $G$}
    \State $C \gets P$
    \For{$i = 1$ to $\lambda$}
        \State Select parents $m_1, m_2$ from $P$
        \State $m_c \gets$ recombine($m_1, m_2$) + Gaussian mutation
        \State $C \gets C \cup \{m_c\}$
    \EndFor
    \ForAll{$m \in C$}
        \State Run simulation with vector field $m(e)$
        \State Render final image $I$
        \State Query VLM with $I$ and prompt $p$ to get fitness score $f$
    \EndFor
    \State $P \gets$ top $\mu$ models in $C$ by score $f$
\EndFor
\State \Return best model from $P$
\end{algorithmic}
\end{algorithm}

\section{Experiments}

\subsection{Experimental Setup}

To evaluate whether meaningful linguistic grounding can emerge from minimal supervision, we trained models using only a single prompt: \textit{``form a cluster''}. The goal was to evolve a P2I that maximizes the P2I-D2R alignment score, as judged by the D2R vision-language model.

To assess the impact of field granularity, we trained separate models at three vector field resolutions: $2 \times 2$, $5 \times 5$, and $10 \times 10$. Each configuration was optimized using a $(\mu + \lambda)$ evolution strategy with $\mu = 5$, $\lambda = 15$, over 50 generations and 30 random seeds. Gaussian mutations ($\sigma = 0.1$) were applied to the P2I model weights.

\subsection{Training Performance on Base Prompt}

\begin{figure}[htbp]
    \centering
    \begin{minipage}[t]{\linewidth}
        \centering
        \includegraphics[width=0.7\linewidth]{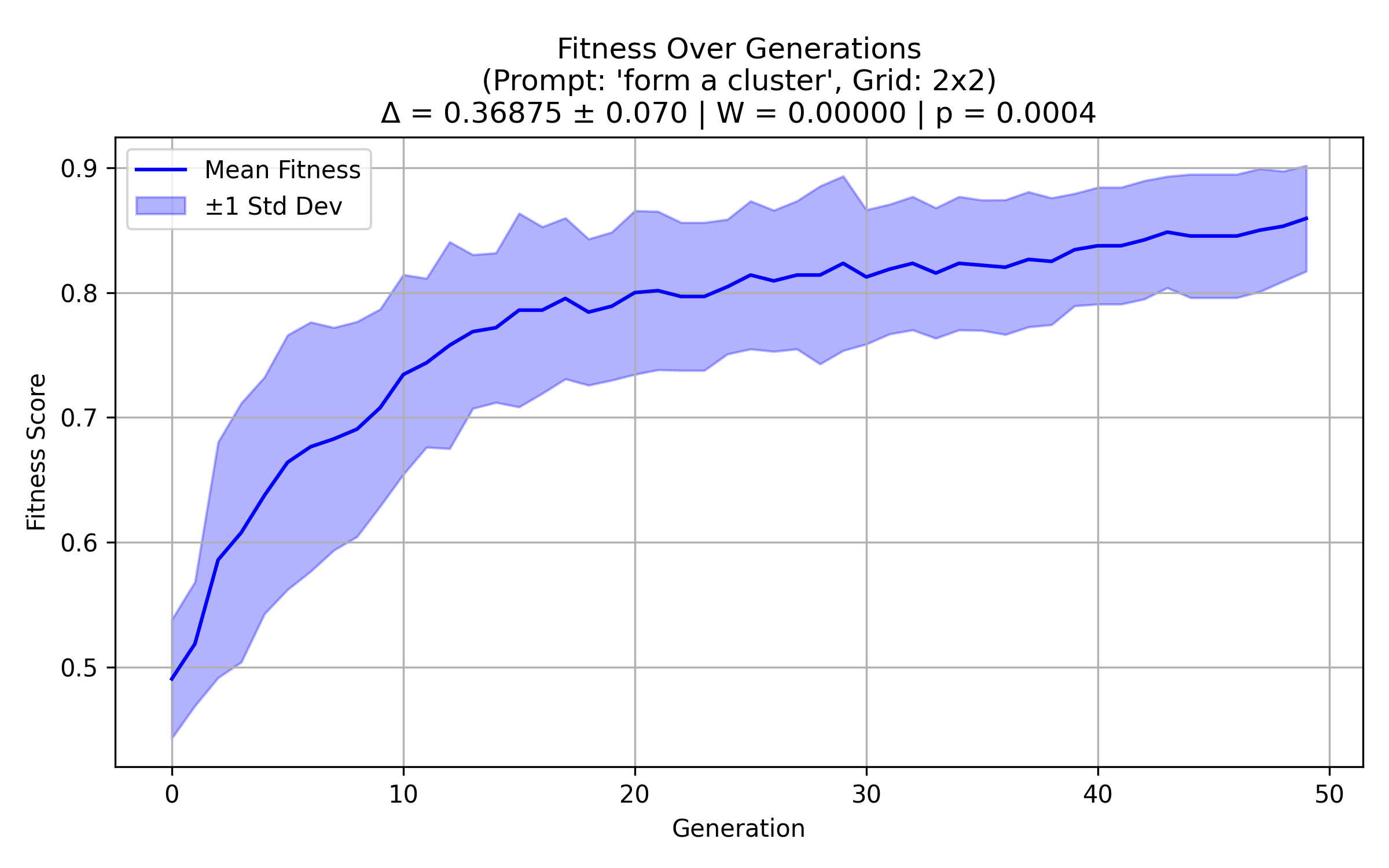}
        \caption*{\footnotesize (a) Grid $2 \times 2$}
    \end{minipage}

    \vspace{0.5em}
    
    \begin{minipage}[t]{\linewidth}
        \centering
        \includegraphics[width=0.7\linewidth]{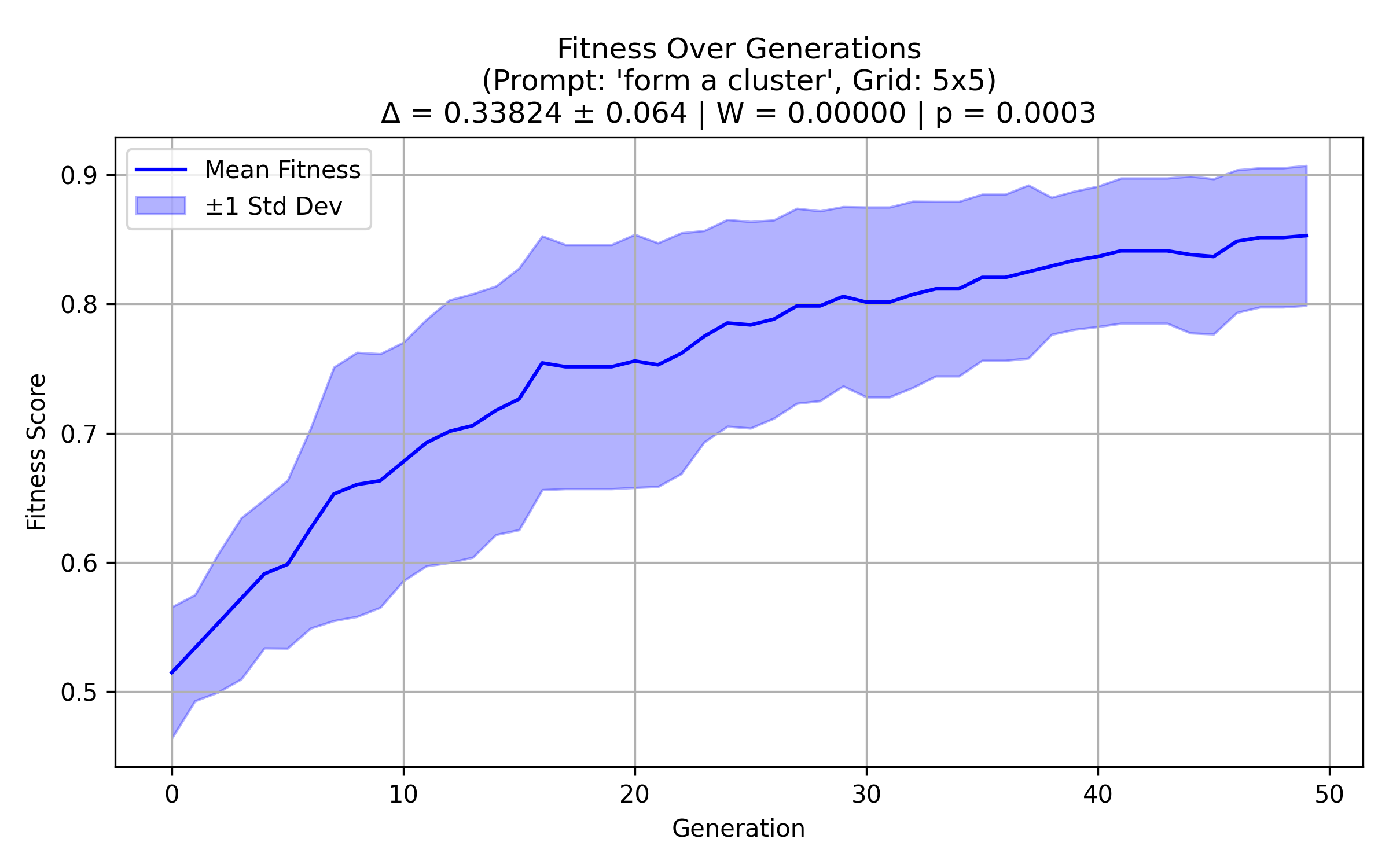}
        \caption*{\footnotesize (b) Grid $5 \times 5$}
    \end{minipage}

    \vspace{0.5em}

    \begin{minipage}[t]{\linewidth}
        \centering
        \includegraphics[width=0.7\linewidth]{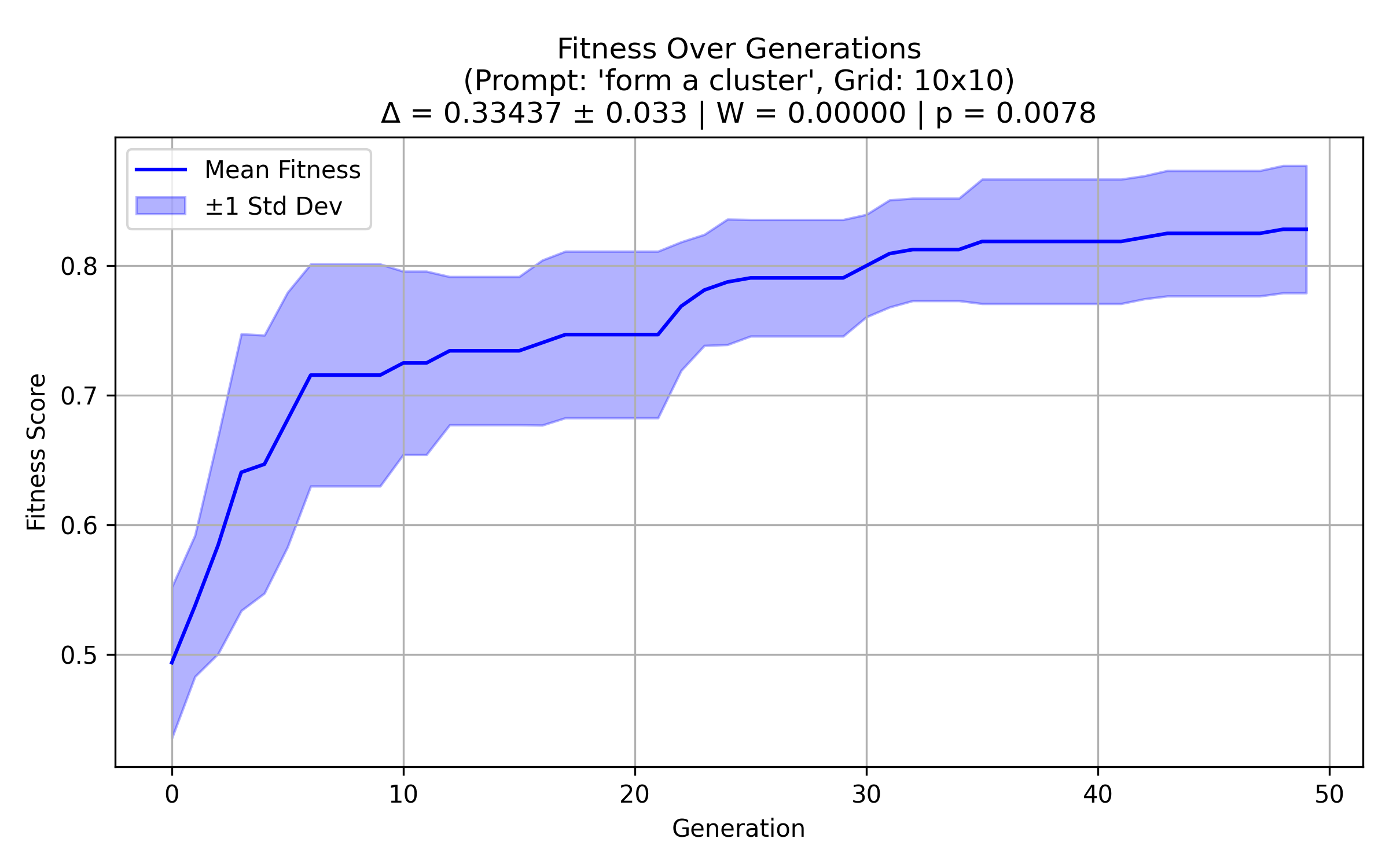}
        \caption*{\footnotesize (c) Grid $10 \times 10$}
    \end{minipage}

    \caption{\footnotesize
        Evolution of fitness over 50 generations for each vector field resolution. Each curve shows mean fitness across seeds; shaded regions indicate standard deviation. All runs used the prompt \textit{``form a cluster''}.
    }
    \label{fig:fitness_curves}
\end{figure}

Figure~\ref{fig:fitness_curves} shows the average P2I-D2R alignment scores over training for each grid resolution. In all cases, fitness improves consistently across generations, demonstrating that the system can evolve effective behaviors using language-based feedback alone.

To assess statistical significance, we performed Wilcoxon signed-rank tests (WSRTs) comparing scores at generation 1 and generation 50. All grid sizes showed significant improvement ($p < 0.01$), confirming that the evolutionary process led to reliable behavioral refinement. Plots include the average gain ($\Delta$), test statistic ($W$), and $p$-value for each setting.

These findings suggest that P2I can align decentralized agent behavior with linguistic instructions, without the need for handcrafted rewards or engineered evaluation criteria.

\subsection{After-Train Quantitative Evaluation}

To assess whether meaningful spatial behavior emerged from vision-language training alone, we conducted a post-hoc quantitative analysis using average pairwise distance (PWD) as a proxy for clustering. For each grid resolution ($2\times2$, $5\times5$, $10\times10$), we selected the best and worst performing seeds based on final VLM alignment scores and ran 30 simulations per seed using the prompt \textit{“form a cluster”}. We measured the change in PWD from initial to final frame and tested for statistical significance using the Wilcoxon signed-rank test.

\begin{figure*}[htbp]
    \centering
    \begin{minipage}[t]{0.28\textwidth}
        \centering
        \includegraphics[width=\linewidth]{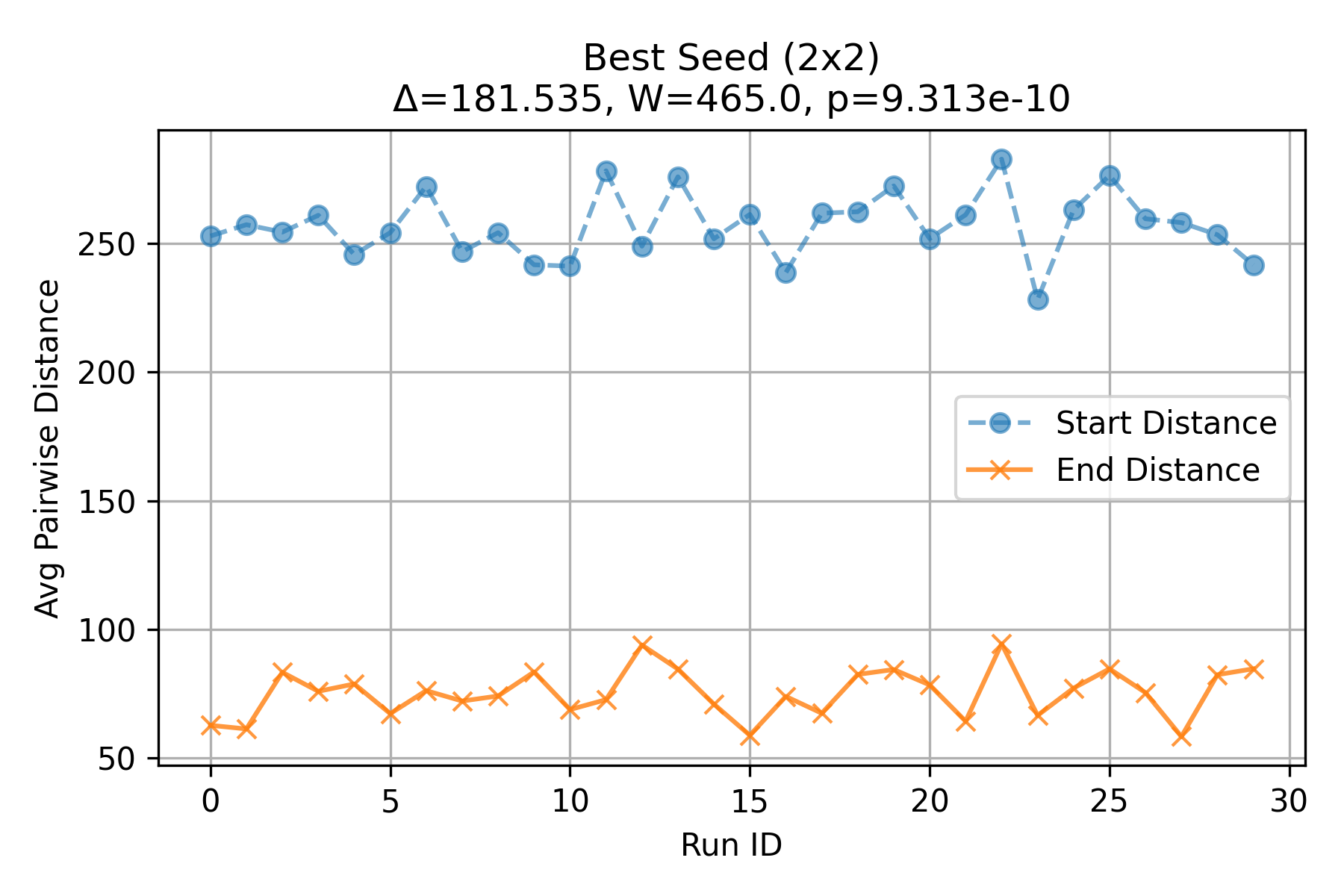}
    \end{minipage}
    \hfill
    \begin{minipage}[t]{0.28\textwidth}
        \centering
        \includegraphics[width=\linewidth]{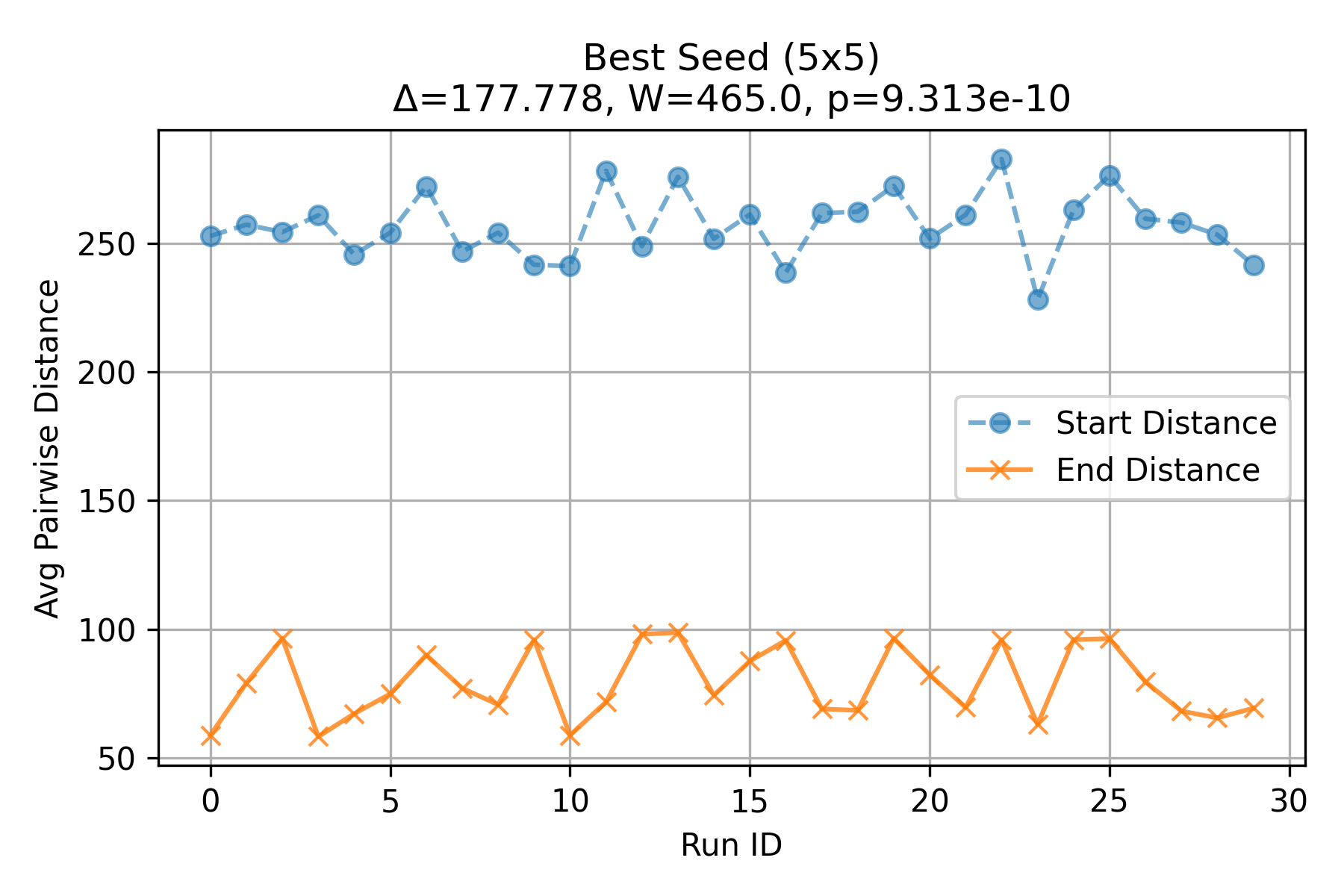}
    \end{minipage}
    \hfill
    \begin{minipage}[t]{0.28\textwidth}
        \centering
        \includegraphics[width=\linewidth]{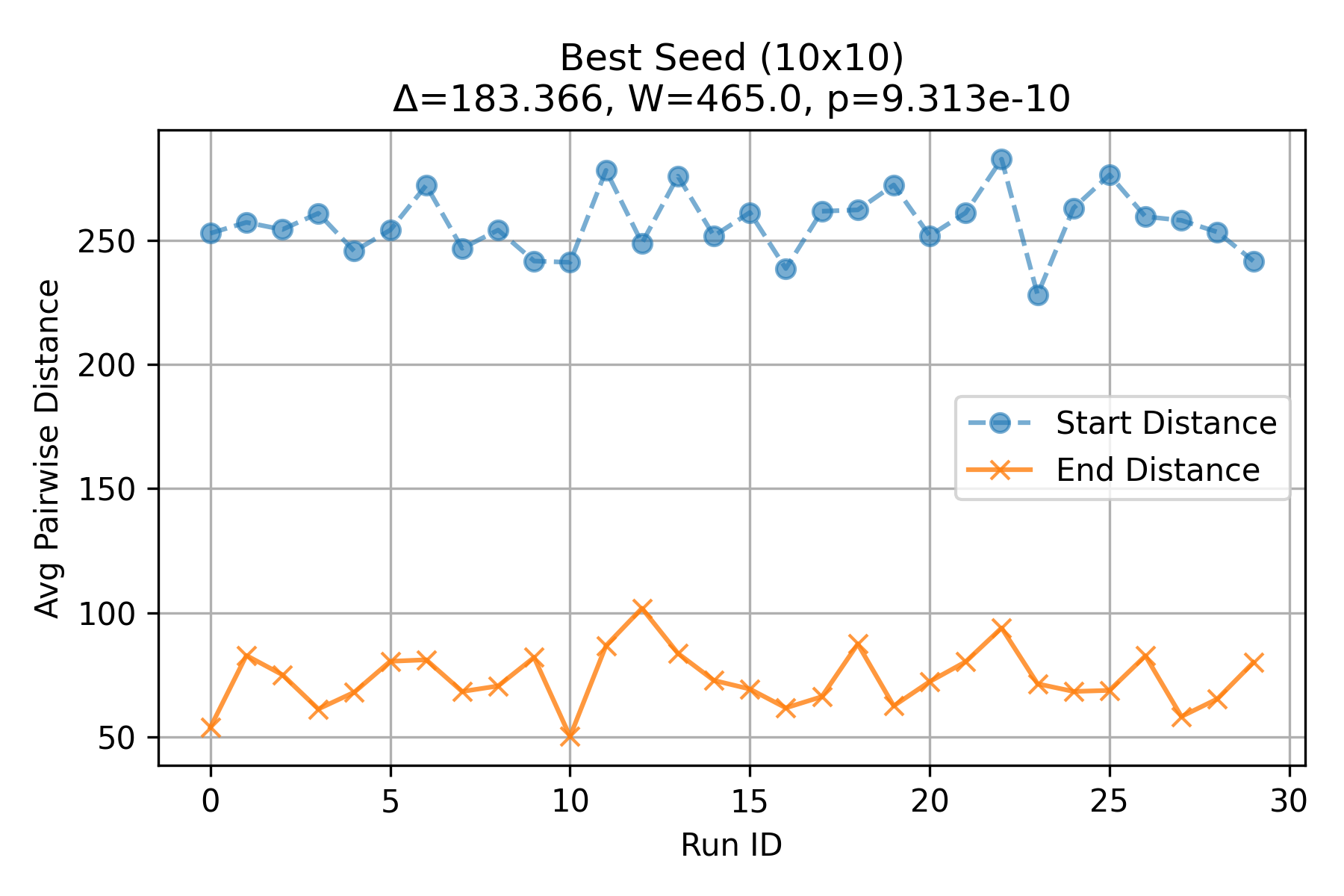}
    \end{minipage}
    
    \vspace{-0.3em}
    
    \begin{minipage}[t]{0.28\textwidth}
        \centering
        \includegraphics[width=\linewidth]{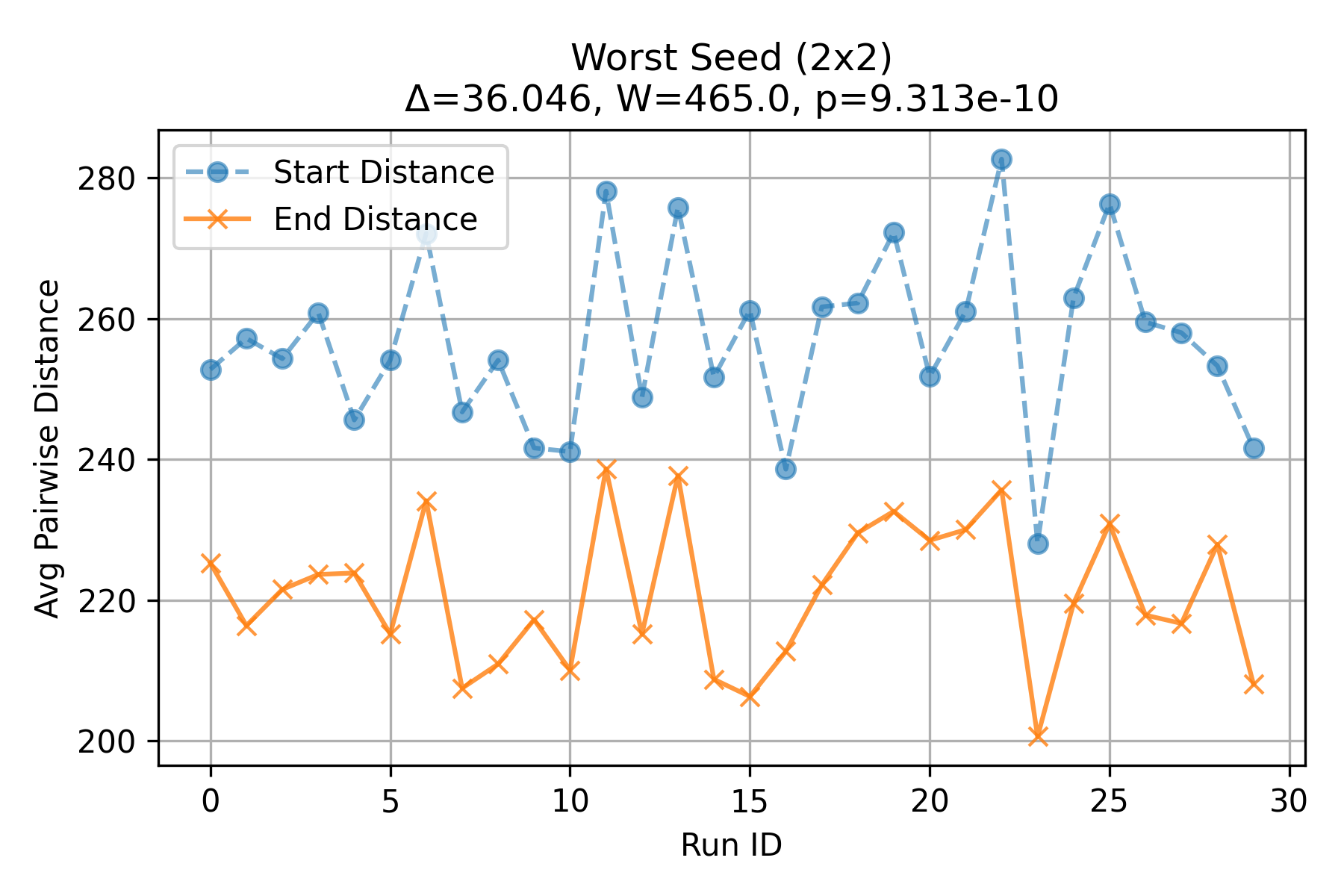}
    \end{minipage}
    \hfill
    \begin{minipage}[t]{0.28\textwidth}
        \centering
        \includegraphics[width=\linewidth]{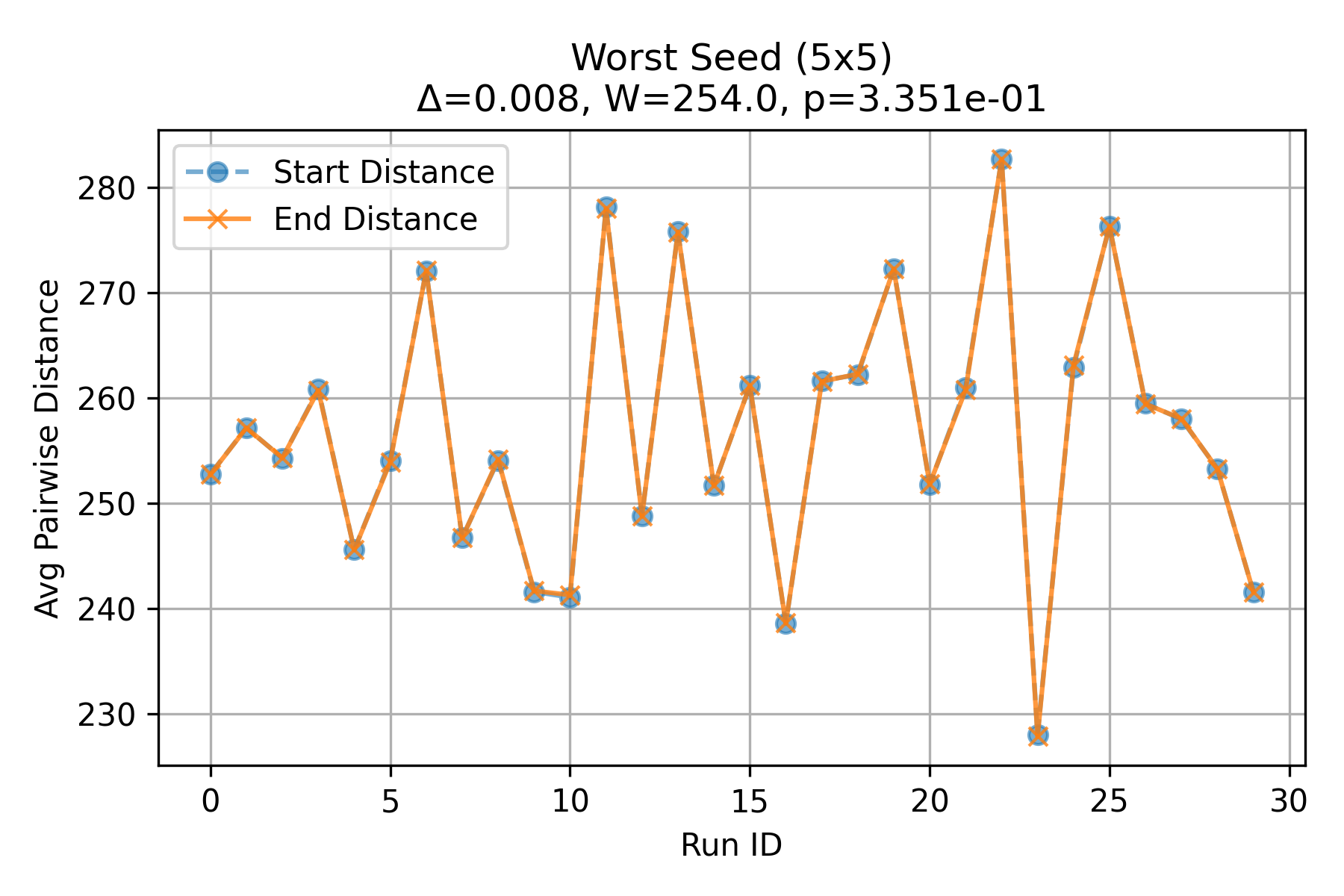}
    \end{minipage}
    \hfill
    \begin{minipage}[t]{0.28\textwidth}
        \centering
        \includegraphics[width=\linewidth]{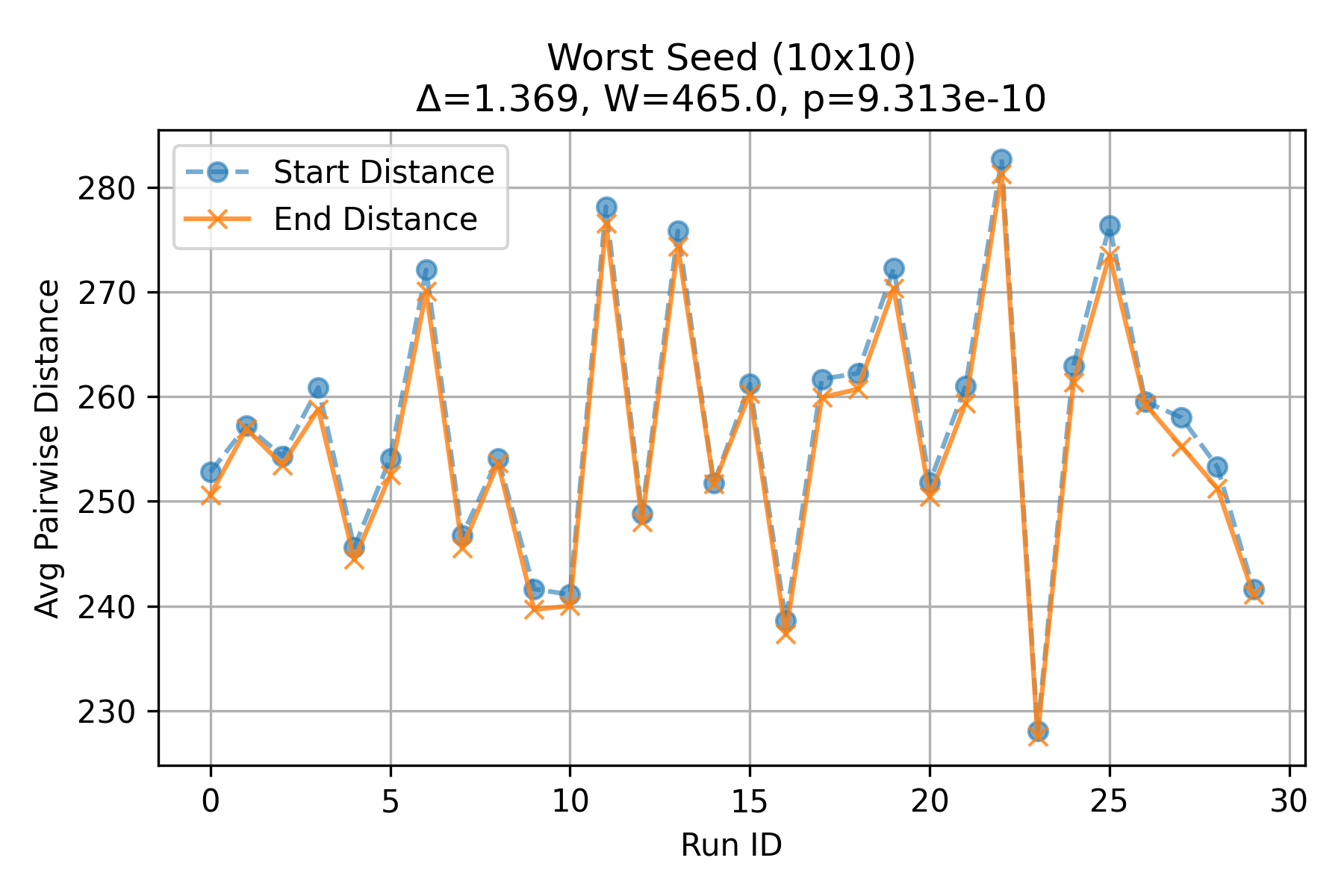}
    \end{minipage}

    \caption{\footnotesize
        Evaluation of clustering behavior using average pairwise distance, comparing best and worst performing seeds for each grid size. Best seeds consistently reduce inter-agent distance after simulation, while worst seeds often fail to induce meaningful change.
    }
    \label{fig:quantitative_eval}
\end{figure*}

As shown in Figure~\ref{fig:quantitative_eval}, the top performing seeds across all grid sizes exhibit significant reductions in PWD ($p < 10^{-3}$), indicating consistent clustering behavior. In contrast, the lowest performing seeds show minimal or inconsistent change—particularly in the $5\times5$ case, where average displacement is near zero and $p > 0.3$.

These findings confirm that the VLM-based alignment score correlates with ground-truth spatial structure, despite never observing it directly. This validates the VLM score as a meaningful training signal and demonstrates that coherent spatial behavior can be evolved using language alone, without handcrafted rewards or spatial supervision.

\subsection{Testing on Unseen Prompts}

To evaluate generalization, we tested the best-performing model (trained on a single prompt: \textit{“form a cluster”}) on eight unseen instructions, including four clustering-related prompts and four with opposite meaning (scattering). All were expressed in free-form language and varied in phrasing, length, and structure (see Table~\ref{tab:prompt_scores}).

The model reliably produced meaningful behaviors in response to clustering prompts, for instance, cells converged into central formations under inputs like \textit{“assemble the cells”}. Importantly, the resulting patterns varied, suggesting the model was not memorizing specific solutions but adapting to linguistic nuances. Even more notably, scattering prompts (e.g., \textit{“drift apart from one another”}) elicited distinct outward movement and dispersion. While the simulation’s boundaries prevented true infinite scattering, cells often spread to corners and edges, forming multiple peripheral clusters. This behavior reflects physical constraints rather than model failure.

\begin{figure*}[htbp]
    \centering
    \includegraphics[width=0.7\textwidth, height=0.5\textheight]{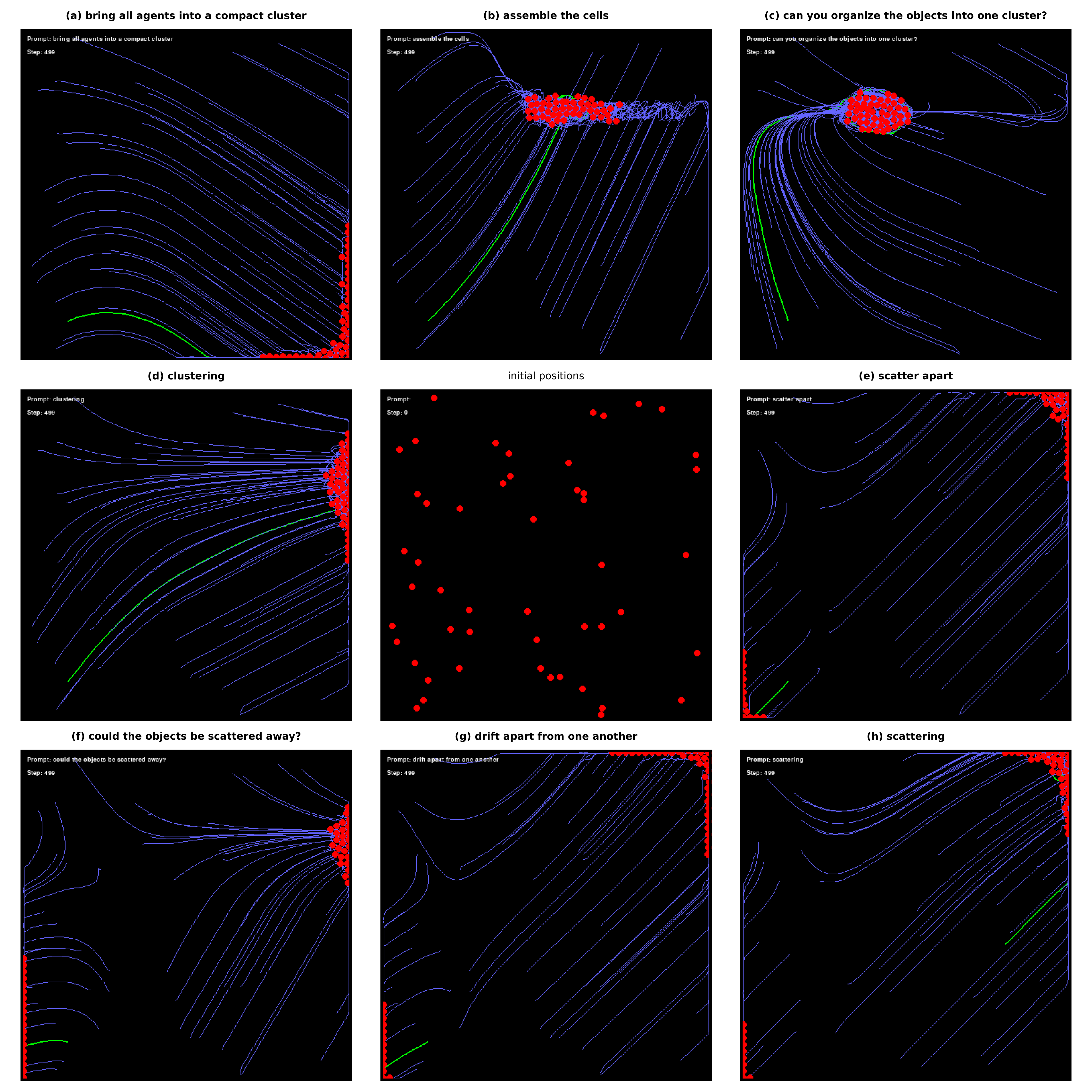}
    \caption{
        Behavioral responses of the trained ZapGPT model to various \textit{free-form natural language prompts}, visualized as final spatial configurations of simulated agents. The \textbf{center panel} shows the \textbf{starting configuration}, while the surrounding panels (a--h) display the \textbf{final positions of agents} under different prompts. Panels (a--d) correspond to \textit{clustering prompts}, directing agents to aggregate, whereas panels (e--h) show the effect of \textit{scattering prompts}, encouraging dispersal. Notably, the model was trained only on the single prompt \texttt{"form a cluster"}, yet it generalizes to a variety of semantically related and contrasting instructions. A video demonstration is available at: \href{https://www.youtube.com/watch?v=u58QZ6P1RUo&list=PLchLYCCaLyQez66FwTCGuUba6-GQrGUXB}{https://www.youtube.com/watch?v=u58QZ6P1RUo}
    }
    \label{fig:test_prompts}
\end{figure*}

Overall, the system demonstrated a clear capacity for semantic inversion and behavioral diversity, despite having seen only one prompt during training. These qualitative results suggest the model supports meaningful generalization to novel linguistic instructions, without retraining or engineered supervision. Example outcomes are shown in Figure~\ref{fig:test_prompts}.

\subsection{Some Quantitative Evaluation}

To support our qualitative findings, we ran 30 simulations per test prompt and evaluated each with the same D2R model used during training. As shown in Table~\ref{tab:prompt_scores}, all eight prompts, in both clustering and scattering categories, achieved alignment scores significantly above 0.5 ($p < 0.01$; one-sided Wilcoxon test), indicating that the model reliably generated behaviors semantically consistent with previously unseen instructions. Interestingly, even prompts with meanings opposite to the training goal (e.g., scattering) elicited coherent, distinct responses. This suggests the system learned an abstract, compositional mapping between language and spatial dynamics, enabling generalization and conceptual inversion beyond the training distribution.

\begin{table}[ht]
\renewcommand{\arraystretch}{1.2}
\centering
\small
\begin{tabular}{|>{\raggedright\arraybackslash}p{4.7cm}|c|c|c|}
\hline
\textbf{Prompt} & \textbf{Mean} & \textbf{Std} & \textbf{p-value} \\
\hline
Assemble the cells & 0.897 & 0.073 & 4.35e$^{-7}$ \\
Bring all agents into a cluster & 0.673 & 0.222 & 7.65e$^{-4}$ \\
Can you organize the objects into one cluster & 0.950 & 0.000 & 2.16e$^{-8}$ \\
Clustering & 0.920 & 0.047 & 2.77e$^{-7}$ \\
Could the objects be scattered away & 0.937 & 0.035 & 9.81e$^{-8}$ \\
Drift apart from one another & 0.927 & 0.063 & 1.30e$^{-7}$ \\
Scatter apart & 0.692 & 0.267 & 6.42e$^{-4}$ \\
Scattering & 0.767 & 0.218 & 7.12e$^{-6}$ \\
\hline
\end{tabular}
\caption{
VLM alignment scores across 30 trials. Top rows: clustering prompts. Bottom: scattering. $p$-values from one-sided Wilcoxon tests vs.\ 0.5.
}
\label{tab:prompt_scores}
\end{table}

\begin{figure}[ht]
\centering
\includegraphics[width=0.7\linewidth]{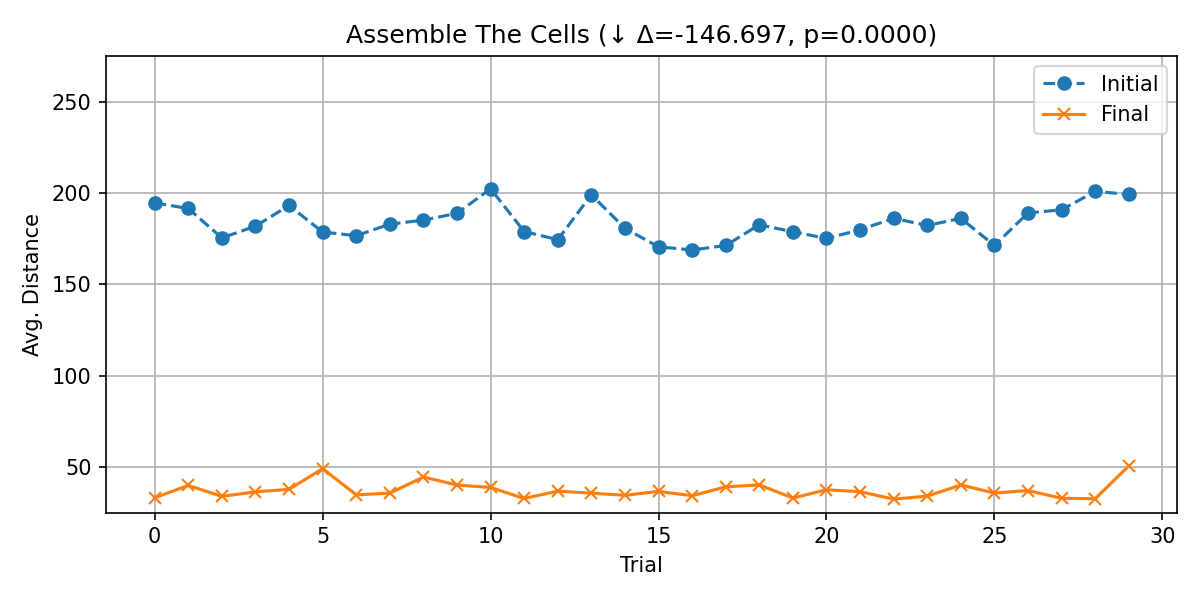}
\vspace{0.5em}
\includegraphics[width=0.7\linewidth]{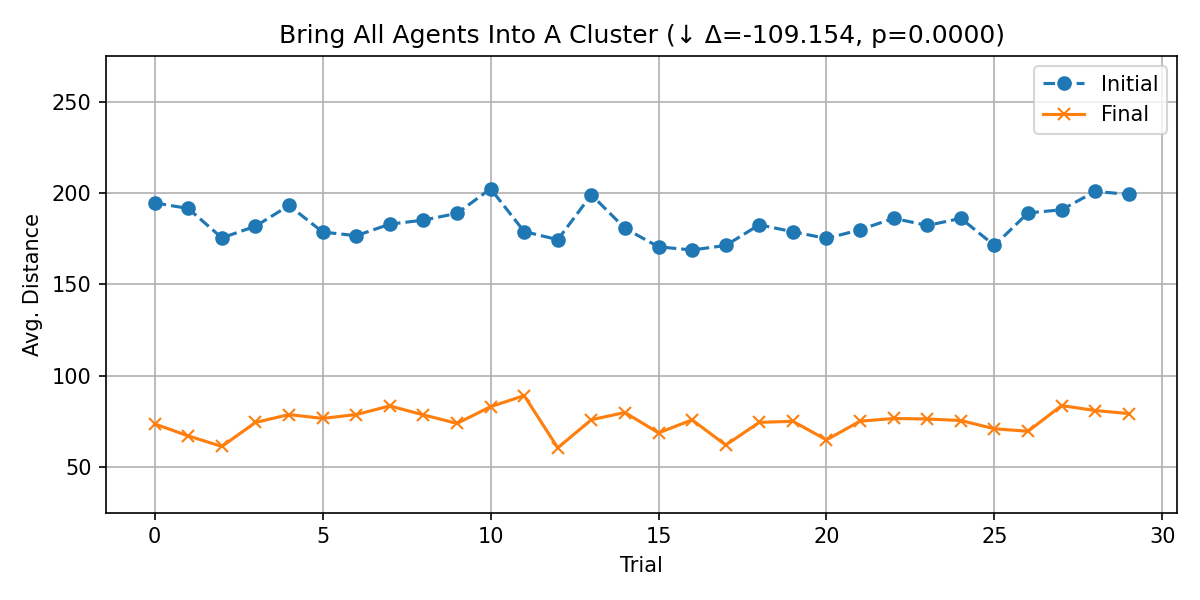}
\vspace{0.5em}
\includegraphics[width=0.7\linewidth]{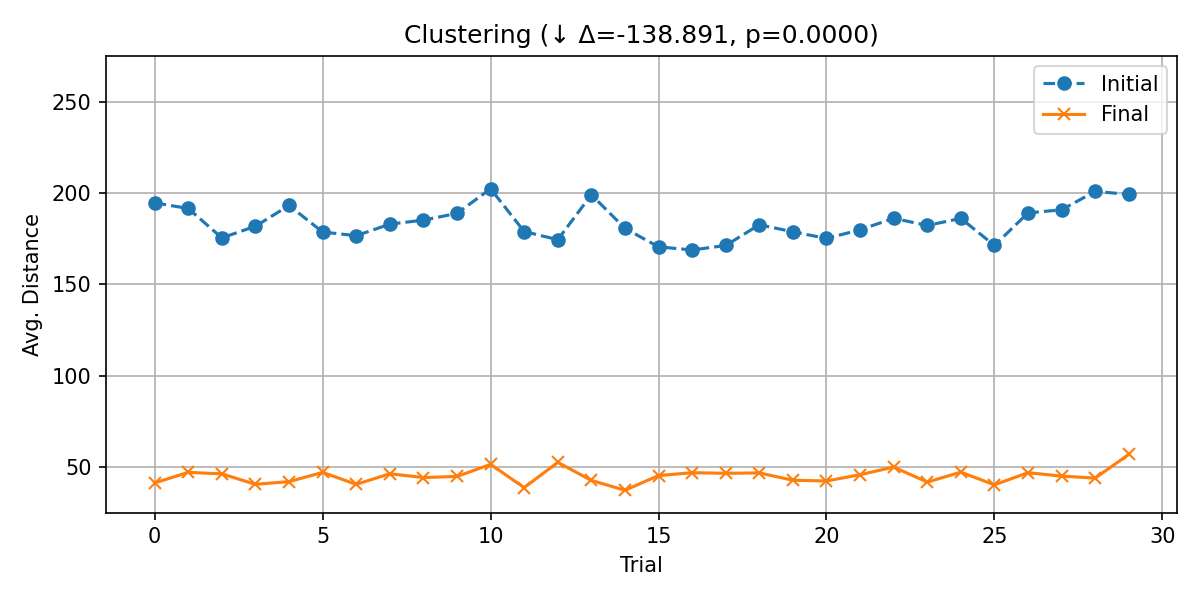}
\vspace{0.5em}
\includegraphics[width=0.7\linewidth]{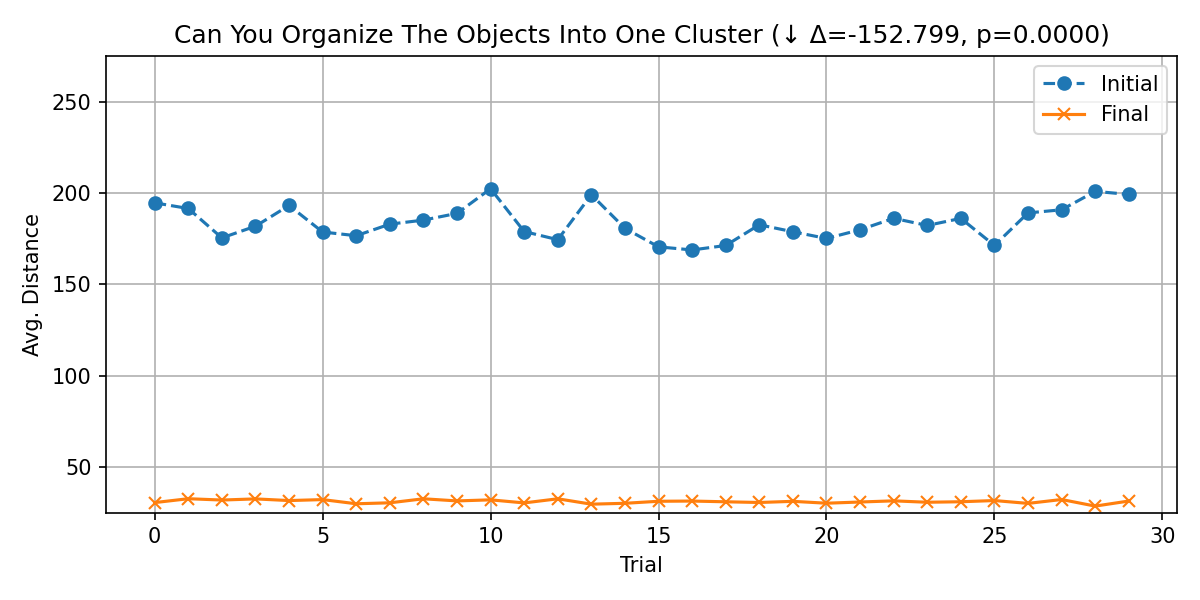}
\caption{Distance change across clustering prompts (30 trials each). Significant reduction confirms successful agent convergence.}
\label{fig:distance_clustering}
\end{figure}

\begin{figure}[ht]
\centering
\includegraphics[width=0.7\linewidth]{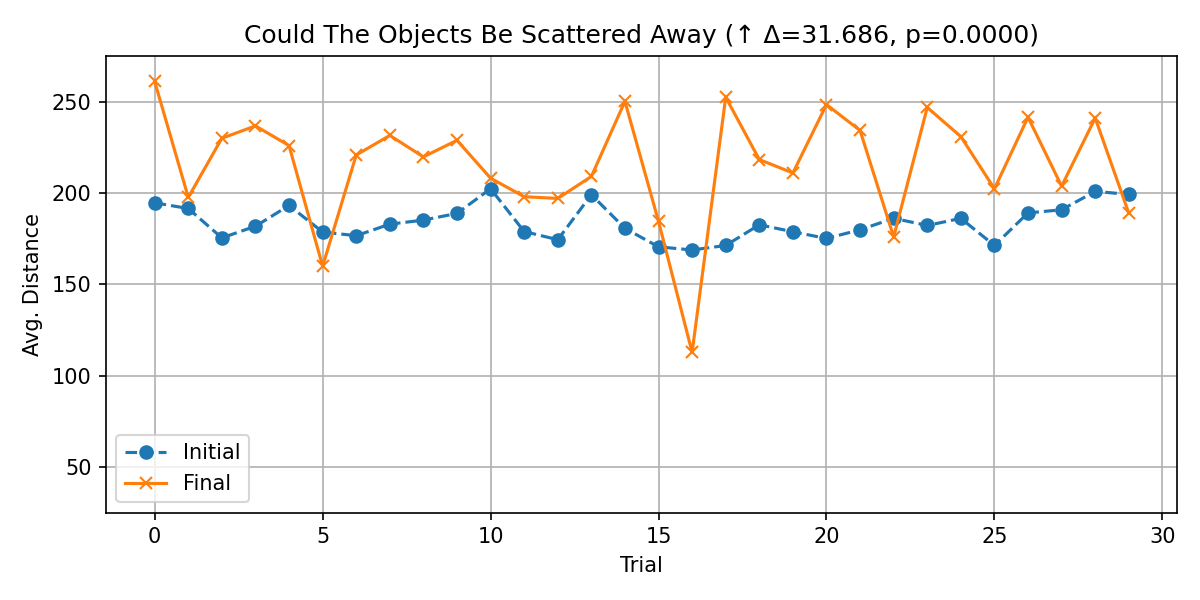}
\vspace{0.5em}
\includegraphics[width=0.7\linewidth]{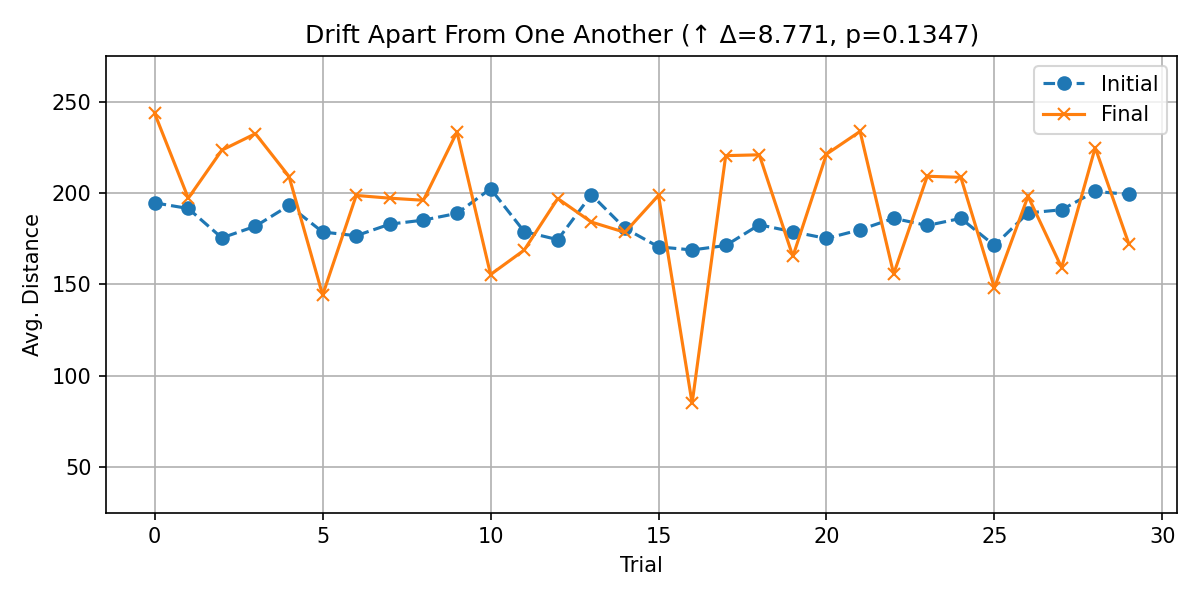}
\vspace{0.5em}
\includegraphics[width=0.7\linewidth]{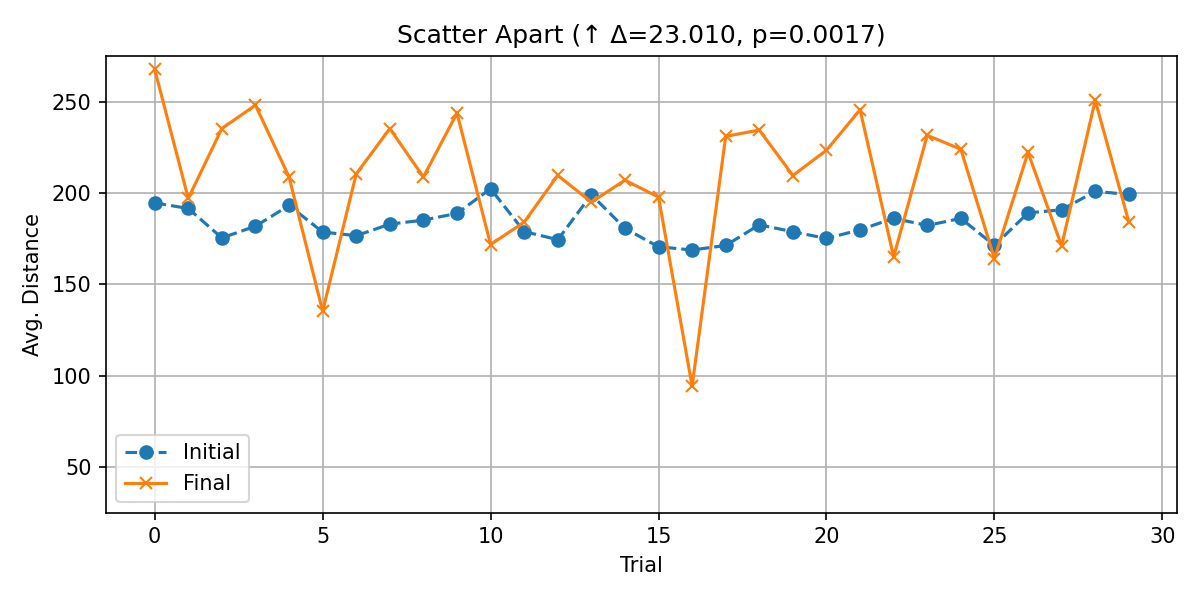}
\vspace{0.5em}
\includegraphics[width=0.7\linewidth]{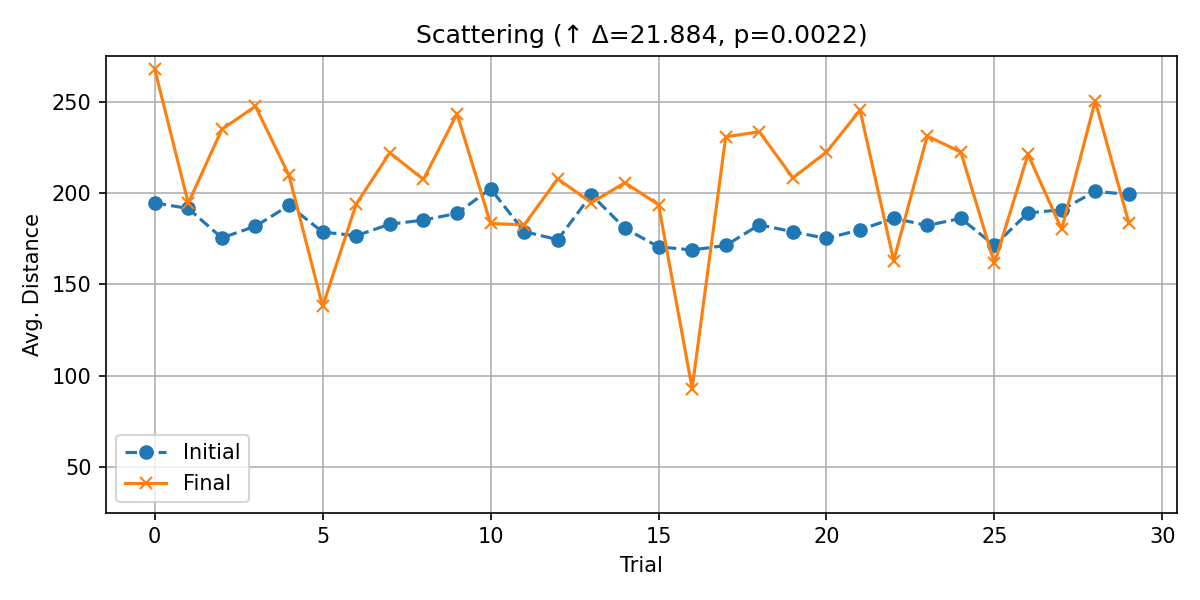}
\caption{Distance change across scattering prompts (30 trials each). Increases confirm dispersal responses.}
\label{fig:distance_scattering}
\end{figure}

As in the training phase, we conducted WSR tests over 30 trials for each prompt, comparing average pairwise distance between the initial and final simulation frames. For clustering prompts (Figure~\ref{fig:distance_clustering}), we observed significant reductions in distance, confirming that agents converged meaningfully. Conversely, scattering prompts (Figure~\ref{fig:distance_scattering}) showed significant increases on 3/4 cases (except for ``Drift apart from one another'' prompt), indicating successful dispersion. These results validate that the learned controller generalizes not only semantically (via VLM scoring), but also spatially, exhibiting consistent physical responses aligned with linguistic intent, even for unseen instructions.

\section{Discussion and Future Work}

\subsection{Summary}

We have presented ZapGPT, a system that translates free-form natural language prompts into spatial vector fields for controlling simulated cellular behavior. A P2I model maps linguistic input to interventions, while a vision-language model (D2R) scores behavioral alignment. The system is trained via evolutionary optimization without engineered rewards or domain-specific supervision. Although trained on a single prompt (``form a cluster''), it generalizes to diverse and even semantically opposite instructions, with quantitative results confirming alignment between linguistic intent and emergent spatial organization.

\subsection{Limitations}

Although the framework shows strong potential, it has notable limitations. First, training was performed on a single prompt (``form a cluster''), limiting linguistic and behavioral diversity. Second, the environment is a simplified, bounded 2D space with non-differentiable physics, which restricts the range of possible behaviors, especially for dispersal tasks, and prevents gradient-based optimization. Incorporating differentiable simulators (e.g., DiffTaichi~\cite{hu2019difftaichi}) could enable end-to-end learning via backpropagation through physical dynamics.

The system also depends on a large, pre-trained vision-language model (Mistral) for behavioral evaluation. This black-box reliance raises concerns about opacity, bias, and computational cost, as every candidate requires a full simulation and VLM query. Future work could reduce overhead by training surrogate evaluators on prompt-image-score tuples or adapting lightweight VLMs to the simulation domain for faster, interpretable scoring.

\subsection{Broader Impact and Applications}

Although demonstrated in a simple simulation, our results suggest that natural language can serve as a high-level, symbolic interface for modulating decentralized, emergent behavior. By treating language as both the source of goals and the medium for evaluation, this framework bypasses engineered metrics and rigid supervision, opening a path toward adaptive, human-aligned control. Such capabilities are relevant not only to artificial life and robotics, but also to bioengineering domains like regenerative medicine, morphogenesis, and synthetic biology, where goals are often ambiguous, dynamic, and best expressed in human terms.

Most prior work has focused on guiding cells in morphospace, shaping their positions or collective movement. However, cells also exist in other high-dimensional spaces: transcriptional (gene expression), metabolic, bioelectric, and physiological states. Many diseases represent misplacements in these spaces, cancer cells occupy the wrong transcriptional attractor, diabetic tissue may be dysregulated metabolically. Language-guided systems could one day help navigate these abstract landscapes, steering cell collectives toward desired state regions without hand-tuned interventions.

Moreover, while our current system receives language as input, future extensions might allow biological or artificial agents to generate language as output. That is, cells or tissues could be placed in closed-loop interaction with language models, using them to signal internal needs or influence their environment. This raises the provocative possibility of giving cells a kind of ``voice,'' enabling them to act through symbolic language rather than just biochemical feedback. From this view, language models become a shared interface: not just tools for humans to program cells, but tools for cells to communicate with humans.

Such a framework blurs the line between interpreter and agent, and invites deeper questions about control, communication, and autonomy in hybrid biological-AI systems. If organs could express needs through language, or if cells could learn to persuade humans to assist in their recovery, this might shift our understanding of agency in living systems, and offer new paradigms for programmable physiology.

\subsection{Zero-Shot Learning and Emergent Generalization}

Although our experimental results arise from a simplified simulation, they demonstrate a degree of generalization: the system was trained on a single prompt (``form a cluster''), yet produced distinct, coherent behaviors when tested on unseen instructions, including both semantically similar and conceptually opposite prompts. These results suggest that the model is not merely memorizing spatial patterns or exploiting grid artifacts. Instead, different prompts give rise to qualitatively different emergent behavior, even without retraining.

This raises important questions about the nature of generalization in such systems. Could this be viewed as a primitive form of zero-shot learning? While we do not make strong claims about true semantic understanding, the observed responses to reversed meaning prompts (e.g., ``scatter apart'') are non-trivial and suggest some level of linguistic modulation over spatial behavior.

Future work will explore this phenomenon more rigorously. One possible direction is to systematically test a broader and more structured set of prompts, varying in lexical similarity, syntactic complexity, and abstractness, and quantify behavioral divergence across them. If consistent behavioral distinctions emerge, this could offer further insight into how linguistic structure maps onto emergent dynamics in decentralized systems.

\subsection{Cognitive Science Analogy} 

An analogy to the P2I–D2R alignment can be drawn from cognitive neuroscience, where symbolic linguistic instructions (e.g., ``reach for the cup'') influence subsymbolic neural activity, dynamically modulating motor systems and sensory processing to generate actions aligned with the symbolic intent. In our framework, symbolic prompts similarly shape subsymbolic processes, spatial vector fields and decentralized agent dynamics, to produce behaviors that match symbolic goals. This mirrors foundational questions in cognitive science around how language is grounded in perception and action, and how meaning can emerge through the interaction of symbolic and embodied systems \cite{harnad1990symbol}.

While we hesitate to claim deep linguistic understanding in such a simple case, the model’s ability to respond non-trivially to reverse-meaning prompts (e.g., ``scatter apart'') suggests an early step toward symbol grounding where linguistic meaning becomes embedded in spatial interventions. This aligns with long-standing questions in cognitive science and AI about how abstract language connects to sensorimotor control and physical representations \cite{harnad1990symbol}.

\subsection{Implications for Artificial Life and Biology}

Artificial life has traditionally focused on understanding how complexity arises from the bottom up: how simple local rules among interacting agents can give rise to lifelike behavior, self-organization, and emergent structure. While this paradigm has yielded profound insights into the nature of decentralized systems, it has historically paid less attention to the role of top-down influences: how global goals, symbols, or high-level constraints can shape local dynamics in return.

ZapGPT is an example of top-down modulation of decentralized processes, mediated through learned mappings and semantic feedback. Understanding this interplay is crucial for ALife, especially in light of major transitions in evolution. When new levels of biological organization emerge, such as multicellularity or social superorganisms, there is often a shift from bottom-up dynamics to hierarchical structure, where top-down regulation begins to guide the behavior and evolution of lower-level units. This bidirectional causality is not an exception but a fundamental feature of living systems.

By enabling symbolic inputs to influence subsymbolic processes, our framework opens new avenues for studying such dynamics in silico. It allows ALife to explore how high-level goals or representations might evolve, stabilize, and exert causal control over emergent system \cite{moreno2015biological}. Ultimately, this helps bridge the gap between emergence and intentionality, offering a richer picture of how living systems, both artificial and natural, integrate top-down and bottom-up dynamics to maintain coherence, adaptability, and evolution.

\textbf{Acknowledgement } The authors wish to acknowledge financial support from Army Research Laboratory contract W911NF2410041.

\bibliographystyle{unsrt}  
\bibliography{references}  


\end{document}